\documentclass[journal]{IEEEtran}
\IEEEoverridecommandlockouts
\usepackage{hhline}
\usepackage{booktabs}
\usepackage{cite}
\usepackage{amsmath,amsthm,amssymb,amsfonts}
\usepackage{array}
\usepackage{graphicx}
\usepackage{textcomp}
\usepackage[table]{xcolor}
\usepackage{xcolor}
\usepackage {hyperref}
\usepackage{soul}
\usepackage{cleveref}
\crefformat{section}{\S#2#1#3}
\usepackage[framemethod=tikz]{mdframed}
\usepackage[ruled, vlined,linesnumbered]{algorithm2e}
\usepackage{algpseudocode}
\usepackage{multirow}
\usepackage{rotating}

\usepackage[font=footnotesize]{caption}
\usepackage{subcaption}
\usepackage{pifont}
\usepackage{longtable}
\usepackage[flushmargin]{footmisc}

\algnewcommand\LeftComment[2]{%
\hspace{#1\algindent}$\triangleright$ \eqparbox{COMMENT}{#2} \hfill %
}

\let\oldnl\nl
\newcommand{\nonl}{\renewcommand{\nl}{\let\nl\oldnl}}

\def\BibTeX{{\rm B\kern-.05em{\sc i\kern-.025em b}\kern-.08em
    T\kern-.1667em\lower.7ex\hbox{E}\kern-.125emX}}
\begin{document}

\title{Approaches and Applications of Early Classification of Time Series: A Review}
\author{\IEEEauthorblockN{Ashish Gupta, Hari Prabhat Gupta, Bhaskar Biswas, and Tanima Dutta}
\vspace{-0.1cm}
 \thanks{The authors are with the Department of Computer Science and Engineering, Indian Institute of Technology (BHU) Varanasi, 221005, India (e-mail: ashishg.rs.cse16@iitbhu.ac.in; hariprabhat.cse@iitbhu.ac.in; bhaskar.cse@iitbhu.ac.in; tanima.cse@iitbhu.ac.in)}}

\maketitle
\begin{abstract}
Early classification of time series has been extensively studied for minimizing class prediction delay in time-sensitive applications such as medical diagnostic and industrial process monitoring. A primary task of an early classification approach is to classify an incomplete time series as soon as possible with some desired level of accuracy. Recent years have witnessed several approaches for early classification of time series. As most approaches have solved the early classification problem using a diverse set of strategies, it becomes very important to make a thorough review of existing solutions. These solutions have demonstrated reasonable performance on a wide range of applications including human activity recognition, gene expression based health diagnostic, and industrial monitoring. In this paper, we present a systematic review of the current literature on early classification approaches for both univariate and multivariate time series. We divide various existing approaches into four exclusive categories based on their proposed solution strategies. The four categories include prefix based, shapelet based, model based, and miscellaneous approaches. We discuss the applications of early classification and provide a quick summary of the current literature with future research directions. 

\end{abstract}

 \noindent \textbf{\textit{Impact statement $-$}} Early classification is mainly an extension of classification with an ability to classify a time series using limited data points. It is true that one can achieve better accuracy if one waits for more data points, but opportunities for early interventions could equally be missed. In a pandemic situation such as COVID-19, early detection of an infected person becomes more desirable to curb the spread of the virus and possibly save lives. Early classification of gas (\textit{e.g.,} methyl isocyanate) leakage can help to avoid life-threatening consequences on human beings. Early classification techniques have been successfully applied to solve many time-critical problems related to medical diagnostic and industrial monitoring. This paper provides a systematic review of the current literature on these early classification approaches for time series data, along with their potential applications. It also suggests some promising directions for further work in this area.

\section{Introduction} Due to the advancement of energy-efficient, small size, and low cost embedded devices, time series data has received an  unprecedented attention in several fields of research, to name a few, healthcare~\cite{liu2015efficient, chen2019automated, 15}, finance~\cite{idrees2019prediction, yin2011financial}, speech and activity recognition~\cite{esling2013multiobjective, pei2017multivariate, 34}, and so on~\cite{aminikhanghahi2018real, 35, shumway2017time}. 
The time series has an inherent temporal dependency among its attributes (data points), which allows the researchers to analyze the behavior of any process over time. Moreover, the time series has a natural property to satisfy human eagerness of visualizing the structure (or shape) of data~\cite{esling2012time}. Numerous algorithms have developed to study various aspects of the time series such as forecasting~\cite{mahalakshmi2016survey}, clustering~\cite{aghabozorgi2015time}, and classification~\cite{bakeoff}. 
The forecasting algorithms attempt to predict future data points of the time series~\cite{mahalakshmi2016survey}. Next, the clustering algorithms aim to partition the unlabeled time series instances into suitable number of groups based on their similarities~\cite{aghabozorgi2015time}. Finally, the classification algorithms attempt to predict the class label of an unlabeled time series by learning a mapping between training instances and their labels~\cite{bakeoff, sharabiani2017efficient}.   

Time Series Classification (TSC) has been a topic of great interest since the availability of labeled dataset repositories such as UCR~\cite{UCR} and UCI~\cite{UCI}. Consequently, a large number of TSC algorithms have emerged by introducing efficient and cutting-edge strategies for distinguishing classes. Authors in~\cite{lines2015time, rakthanmanon2013addressing, sharabiani2017efficient} focused on instance based learning where the class label of a testing time series is predicted based on a similarity measure. Dynamic Time Warping (DTW)~\cite{berndt1994using} and its variations~\cite{rakthanmanon2013addressing, sharabiani2017efficient} with $1$-Nearest Neighbors (1-NN) have been extensively used similarity measures in the instance based TSC algorithms.

Recently, deep learning based TSC algorithms, discussed in~\cite{fawaz2019deep}, have also demonstrated a significant progress in time series classification. Two robust TSC algorithms are proposed in~\cite{wang2017time} and~\cite{liu2018time}, by using ResNet and Convolutional Neural Network (CNN) framework, respectively. The authors in~\cite{9127499} developed a reservoir computing approach for generating a new representation of Multivariate Time Series (MTS). The approach is incorporated into recurrent neural networks to avoid computational cost of the back propagation during classification. In~\cite{karim2019multivariate}, the authors proposed a multivariate TSC approach by combining two deep learning models, Long Short-Term Memory (LSTM) and Fully Convolutional Network (FCN), with an attention mechanism. Two recent studies~\cite{yoon2019time, esteban2017real} employed generative adversarial networks for TSC by modeling the temporal dynamics of the data.

The main objective of TSC algorithms is to maximize the accuracy of the classifier using complete time series. However, in time-sensitive applications such as gas leakage detection~\cite{8}, earthquake~\cite{Fauvel2020ADM}, and electricity demand prediction~\cite{dachraoui2013early}, it is desirable to maximize the earliness by classifying an incomplete time series. A classification approach that aims to classify an incomplete time series is referred as early classification~\cite{1,2,4}. Xing \textit{et al.}~\cite{1} stated that the earliness can only be achieved at the cost of accuracy. They indicated that the main challenge before an early classification approach is to optimize the balance between two conflicting objectives, \textit{i.e.}, accuracy and earliness. One of the first known approaches for early classification of time series is proposed in~\cite{3}, and then after several researchers have put their efforts in this direction and published a large number of research articles at renowned venues. After doing an exhaustive search, we found a minor survey in~\cite{santos2016literature}, which included only a handful of existing early classification approaches and did not provide any categorization.

This paper presents a systematic review of the early classification approaches for both univariate and multivariate time series data. At first, we discuss the potential applications that motivated the researchers to work in the early classification of time series. Section~\ref{applications} provides the detail of the applications. In Section~\ref{method}, we explain the research methodology for searching, filtering, selection of the reviewed papers. Section~\ref{fundamental} discusses the fundamentals of early classification approaches and their categorization. We categorize the approaches into four groups based on the solution strategies that the researchers adopted for early classification. The included approaches in the proposed categories are detailed in four subsequent sections. Finally, Section~\ref{dis} summarizes the review by discussing the challenges of the solution approaches and their recommendations for future work.

\section{Applications of Early Classification}\label{applications}
In data mining and machine learning, early classification of time series has received significant attention as it can solve time-critical problems in many areas including medical, industry, and transportation. Literature indicates numerous applications of early classification of time series. Some of the important applications are illustrated in Fig.~\ref{applications} and also discussed in detail as follows.
\begin{figure}[h]
\centering
\includegraphics[scale=1]{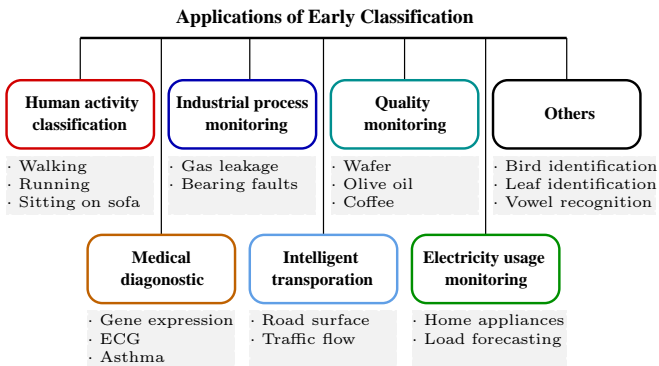}
\caption{Applications of the early classification of time series.}
\label{applications}
\end{figure}
\subsubsection{\textbf{Human activity classification}} With the availability of multi-modal sensors in smartphones and wearable, people can easily monitor their daily routine activities such as walking, running, eating, and so on. 
The early classification of human activities helps to minimize the response time of the system and in turn, improves the user experience~\cite{34}. The researchers in~\cite{34} attempted to classify various complex human activities such as sitting on sofa, sitting on floor, standing while talking, walking upstairs, and eating, using the sensors generated MTS. The studies in~\cite{16, 21, 33} focused on identifying human actions such as pick up, chicken dance, golf swing, \textit{etc.}, using an incomplete MTS of motion angles. The authors in~\cite{44} differentiated between the normal and dangerous human behavior by classifying the sequential patterns of the activities such as walking followed by lying down or falling. In the studies~\cite{37, 43}, the authors have attempted to classify $19$ different human activities using only partial time series.     
\subsubsection{\textbf{Medical diagnostic}} A primary motivation of the work in~\cite{11, 12, 14, 25, 13, 15, 17,19, 20, 24, 25} is to develop the early classification approaches for medical diagnosis of the diseases such as asthma, viral infection, abnormal Electrocardiogram (ECG), \textit{etc.} Early diagnosis of these diseases can significantly minimize the consequences on patient health and assist the doctors in treatment. Gene expression has been used to study the viral infection on patients, drug response on the disease, and patient recovery~\cite{11, 12, 14}. 
Early detection of asthma can help to prevent life-threatening risk and further to provide rapid relief~\cite{19}. The study in~\cite{24} focused on predicting the right time for transferring the patient to Intensive Care Unit (ICU), using the MTS of physiological measurements such as temperature and respiratory rate. Further, ECG is also a time series of electrical signals that are generated from heart's activity. 
Early classification of ECG~\cite{25,13, 15, 38} helps to diagnose abnormal heart beating at the earliest, reducing the risk of heart failure. 

\subsubsection{\textbf{Industrial process monitoring}} With the advancements in sensor technology, monitoring the industrial processes has become convenient and effortless by using the sensors. The sensors generate time series, which is to be classified for knowing the status of the operation. The authors in~\cite{8, 53, 43, 3, 28, 30, 46} are motivated to build early classification based solutions for industrial problems by using sensory data. In chemical industries, even a minor leakage can cause hazardous effects on the crew members' health~\cite{8}. Early classification not only reduces the risks of health but also minimizes the maintenance cost by ensuring smooth operations all the time. In particular, an electronic nose is developed using gas sensors in~\cite{8}, to smell the gas odor. It generates an MTS, which needs to be classified as early as possible to detect any leakage. 
In~\cite{43}, the authors attempted to detect the problems such as pump leakage, reduced pressure, and inefficient operation, in a hydraulic system. Early detection of these problems can significantly lower the maintenance cost. Early identification of instrumentation failure in nuclear power plants can save hazardous consequences~\cite{3, 27, 46}.     
\subsubsection{\textbf{Intelligent transportation}} As modern vehicles are equipped with several sensors, it becomes easy to monitor the behavior of driver, road surface condition, and traffic flow prediction, by using the generated sensory data. If the driver is overspeeding or alcoholic, early classification of driving pattern can reduce the chances of an accident that may occur due to delayed classification. 
The study~\cite{35} attempted to early classify the type of road surface by using the sensors such as accelerometer, light, temperature, \textit{etc.} Such early classification of the road surfaces helps to choose an alternative path if the surface condition is poor, \textit{e.g.,} bumpy or rough. The studies in~\cite{16, 43} focused on identifying weekdays using traffic flow MTS. It helps to forecast the road traffic for that particular day.   
\subsubsection{\textbf{Quality monitoring}} It is a process of ensuring product quality based on preset standards. With the help of sensors or spectroscopy, the quality of a product can be ensured during its manufacturing. Early detection of low-quality product is desirable to avoid negative consequences on the production. In semiconductor industries, checking the quality of silicon wafers is a crucial task. Early classification of wafer quality~\cite{20, 13, 15, 17, 19, 38, 22, 29} can help to minimize the maintenance cost while ensuring the smooth operations all the time. The Origin of olive oils can be distinguished geographically using time series data obtains from spectroscopy~\cite{10, 32}. The work in~\cite{47, 30} attempted to classify two types of coffee beans, \textit{i.e.,} Robusta and Aribica. Further, the quality of beef can also be ensured using the spectroscopy~\cite{46}.
\subsubsection{\textbf{Electricity usage monitoring}} Awareness about electricity usage helps the consumers avoid unnecessary wastage of the energy and curb the monthly electricity bill. Early classification of the currently running appliances can reduce usage by turning them off during peak hours.  The work in~\cite{33} has successfully classified several household appliances including air conditioner, washing machine, microwave oven, and so on. The studies in~\cite{28, 46, 29, 6} have been utilizing the electricity usage data for distinguishing the seasons from April to September and from October to March. 
\subsubsection{\textbf{Others}}  
The authors in~\cite{29} conducted a case study to identify a bird by using only $20\%$ of time series obtains from chirping sound. The work in~\cite{36} distinguished major Indian rivers based on the time series of water quality parameters. Early prediction of the stock market (IBEX 35) can help to plan better strategies for investment~\cite{47}. In addition, the early classification has shown good performance on Japanese vowel recognition~\cite{38, 26}, Australian sign language~\cite{3, 22, 16}, and leaf identification~\cite{27,47, 30}. 

\vspace{0.5cm}
\section{Research protocol}\label{method}
In order to conduct a systematic review of early classification approaches for time series, we followed a review style similar to~\cite{albahri2020role, albahri2018systematic} and developed a research protocol for searching, filtering, and selection of included reviewed paper. The protocol consists of the following steps: 
\begin{itemize}
 \item \textit{Search strategy:} To search the papers from standard databases using relevant queries. 
 \item \textit{Inclusion criteria:} To filter out the related papers. 
 \item \textit{Selection process:} To select the final set of papers included in this review. 
 \end{itemize}
This review gives a quick understanding of the notable contributions that have been made over the years in the area of early classification of time series. 
\vspace{0.3cm}
\subsubsection{\textbf{Search strategy}}
At first, we form query terms that can fetch most of the relevant papers from the standard databases. For this review, we selected following databases: IEEE Xplore (IX), Science Direct (SD), and Google Scholar (GS). These databases cover almost every aspect of research in the engineering field including computer science and biomedical.  
Table~\ref{query} shows different query terms and the number of papers fetched from the selected databases. As we intended to find maximum number of papers that have studied early classification of time series, we formed three different search queries at varying level of granularity while keeping the term \textit{early} in common. The search results include the papers from last $20$ years (\textit{i.e.,} from $2000$ to $2020$).     

\begin{table}[h]
\centering
\caption{Search query terms and the papers fetched from the databases.}
\begin{tabular}{|c|c|c|c|}
\hline
        & \textbf{Query terms (in title)}                                                                                                                                                              & \textbf{\begin{tabular}[c]{@{}c@{}}Fetched \\ papers\end{tabular}}                                                                                    & \textbf{\begin{tabular}[c]{@{}c@{}}Final \\ papers\end{tabular}}    \\ \hline
Query 1 & \begin{tabular}[c]{@{}c@{}}``early classification" AND \\ ``time series" \end{tabular}                                                                                                                                         & \begin{tabular}[c]{@{}c@{}}IX = 14\\  SD = 02\\  GS = 53\end{tabular} & \begin{tabular}[c]{@{}c@{}} \textbf{69} \\ (19 duplicate)\end{tabular}     \\ \hline
Query 2 & \begin{tabular}[c]{@{}c@{}}\big (``early detection" OR \\``early prediction" OR \\ ``early recognition" \big) AND\\  \big(``ongoing" OR  \\ ``time series" OR ``sequence"\big)\end{tabular}                        & \begin{tabular}[c]{@{}c@{}}IX = 13\\  SD = 13\\ GS = 90\end{tabular} & \begin{tabular}[c]{@{}c@{}} \textbf{116} \\ (44 duplicate)\end{tabular}\\ \hline
Query 3 & \begin{tabular}[c]{@{}c@{}}\big(``early detection" OR \\ ``early prediction" OR \\``early recognition" OR \\``early classification"\big) AND \\ \big(``temporal patterns" OR \\ ``observations"\big)\end{tabular} & \begin{tabular}[c]{@{}c@{}}IX = 06\\ SD = 01\\ GS = 21\end{tabular} & \begin{tabular}[c]{@{}c@{}} \textbf{28} \\ (08 duplicate)\end{tabular}  \\ \hline
\multicolumn{1}{|l}{}         & \multicolumn{1}{l}{}                                                                                                                                                                                   & \multicolumn{1}{c|}{  \textbf{Total}}                                                                                    & \multicolumn{1}{c|}{\textbf{213} }    \\ \hline
\end{tabular}
\label{query}
\end{table}
\subsubsection{\textbf{Inclusion criteria}}
A paper is included in this review only if it meets the following criteria:
\begin{itemize}
 \item It must be written in English only. 
 \item It  should be a book chapter, conference, or journal paper.
 \item The early classification approach in the paper should essentially be developed for time series data only. 
 \item The data points in the time series should be of numeric type. However, it may be a transformed version of any other type of data (\textit{e.g.,} image or video).
\end{itemize}
\vspace{0.3cm}
\subsubsection{\textbf{Selection process}}
\begin{figure}
\centering
\includegraphics[scale=1]{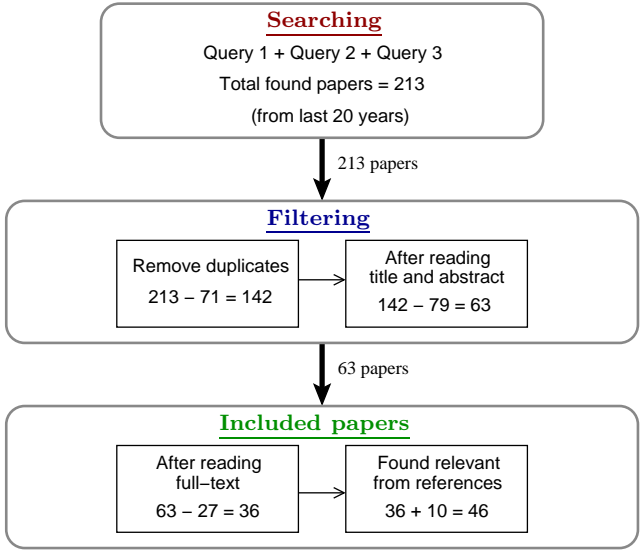}
\caption{Illustration of the paper selection process.}
\label{methodology}
\end{figure}

Once the papers are fetched using three aforementioned query terms, we first remove the duplicate papers and then review the title and abstract of the remaining. After reading the title and abstract, several papers are filtered out as they could not meet the inclusion criteria. Later, rest of the papers were subjected to full-text reading and some of them are also removed at this stage. Finally, we also include some relevant papers from the references. Fig.~\ref{methodology} illustrates the selection process with the number of filtered papers at different stages. 

\section{Fundamentals and Categorization of Early Classification Approaches}\label{fundamental}
In this section, we first discuss the fundamentals of time series prerequisite for acquiring a sound understanding of various early classification approaches. Later, we present the categorization of the approaches.
\subsection{Fundamentals}
This subsection defines the terminologies and notations used in this paper.          
\subsubsection{\textbf{Time series}} It is defined as a sequence of $T$ ordered observations typically taken at equal-spaced time intervals~\cite{ye2009time}, where $T$ denotes the length of complete time series. A time series is denoted as $\mathbf{X}^d = \{X_1, X_2, \cdots, X_T\}$, where $d$ is the dimension and $X_i \in \mathbb{R}^d$ for $1 \leq i \leq T$. If $d=1$ then the time series is referred as univariate otherwise multivariate. If the time series is a dimension of MTS then it can be referred as component~\cite{34,35}. In general, a time series is \textit{univariate} unless it is explicitly mentioned as \textit{multivariate}.                
\subsubsection{\textbf{Time series classification}} It refers to predicting the class label of a time series by constructing a classifier using a labeled training dataset~\cite{bakeoff}. Let $D$ is a training dataset consisting $N$ instances as $N$ pairs of time series $\mathbf{X}$ and their class labels $y$. The time series classifier learns a mapping function $\mathcal{H}: \mathbf{X} \rightarrow y$. The classifier can predict the class label of a testing time series $\mathbf{X}^{'} \notin D$ only if it is complete, \textit{i.e.,} the length of $\mathbf{X}{'}$ should be the same as that of training instances~\cite{5}. 
                
\subsubsection{\textbf{Early classification of time series}} 
\begin{figure}[h]
\centering
\includegraphics[scale=1]{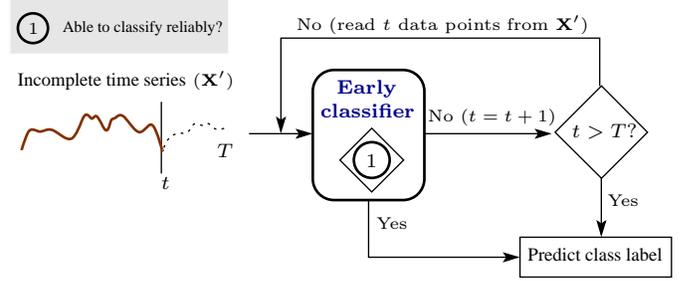}
\caption{Illustration of an early classification framework for time series.}
\label{ec}
\end{figure}

According to~\cite{30}, early classification is an extension of the traditional classification with the ability to classify an unlabeled incomplete time series. In other words, an early classifier is able to classify a testing time series with $t$ data points only, where $t \leq T$. Early classification is desirable in the applications where data collection is costly or late prediction causes hazardous consequences~\cite{35}. Intuitively, an early classifier may take more informed decision about class label if more data points are available in the testing time series~\cite{6} but it will delay the decision. Therefore, the researchers focused on optimizing the accuracy of prediction with minimum delay (or maximum earliness). Further, the early classification of time series is analogous to a case of missing features with the constraint that the features are missing only because of unavailability of data points~\cite{6}. 
In the context of early classification, a testing time series can be referred as incomplete or incoming.   
Fig.~\ref{ec} illustrates an early classification framework for predicting the class label of an incomplete time series $\mathbf{X}{'}$.  

\begin{figure*}
\centering
\includegraphics[scale=1]{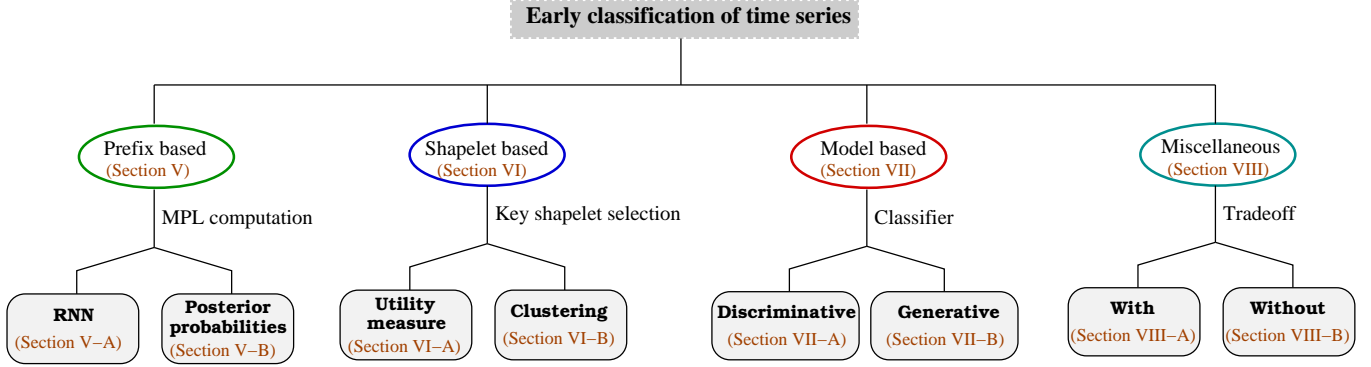}
\caption{Categorization of the early classification approaches. [MPL: Minimum Prediction Length, RNN: Reverse Nearest Neighbor]}
\label{catect}
\end{figure*}

\subsubsection{\textbf{Earliness}} It is an important measure to evaluate the effectiveness of the early classifiers. Let the early classifier uses $t$ data points of a testing time series during classification. Now, the earliness is defined as $\mathbf{E} = \frac{T-t}{T} \times 100$, where $T$ is the length of complete time series~\cite{34}. The earliness is also called as timeliness~\cite{7}.  
\subsubsection{\textbf{Prefix}} In~\cite{2}, prefix of a time series $\mathbf{X}$ is given as the following subsequence $\mathbf{X}[1,t] = \mathbf{X}[1], \mathbf{X}[2], \cdots, \mathbf{X}[t]$, where $t$ denotes the length of the prefix. The training dataset is said to be in prefix space if it contains only the prefix of the time series with their associate class labels.  
       
\subsubsection{\textbf{Shapelet}} It is defined as a quadruple $S = (s, l, \delta, y)$, where $s$ is a subsequence of length $l$, $\delta$ is the distance threshold, and $y$ is the associated class label~\cite{10, 11}. The distance threshold $\delta$ is usually learned using the training instances and it is used to find whether the shapelet is matched with any subsequence of the testing time series.       
\subsubsection{\textbf{Interpretability}} It mainly refers to the fact that how convincing the classification results are to the domain experts. In the medical applications, adaptability of any early classification approach heavily relies on its interpretability~\cite{4}. The authors in~\cite{4, 10, 14, 15} assert that a short segment of the time series is more convincing and helpful than the time series itself if such a segment contains class discriminatory patterns.      
\subsubsection{\textbf{Reliability}} 
It expresses the guarantee that the probability of early predicted class label of an incomplete time series is met with a user-specified threshold~\cite{6, 7}. Reliability is a crucial parameter to ensure minimum required accuracy in the early classification. It is also termed as uncertainty estimate or confidence measure in different studies~\cite{10, 29}.    

\subsection{Categorization of early classification approaches}
This work categorizes the early classification approaches (discussed in the selected papers) into meaningful groups, to better understand their differences and similarities. We believe that one of the most meaningful ways to categorize these approaches is the strategies that they have discovered to achieve the earliness. We broadly categorize the approaches into four major groups, as shown in Fig.~\ref{catect}. The summary of the included papers in different groups is given in Table~\ref{summary1}.
\subsubsection{\textbf{Prefix based early classification}}
The strategy is to learn a minimum prefix length of the time series using the training instances and utilize it to classify a testing time series. During training, a set of $T$ classifiers, one for each prefix space, are constructed. The classifier that achieves a desired level of stability with minimum prefix length, is considered as early classifier and the corresponding prefix length is called as Minimum Prediction Length (MPL)~\cite{2, 5, 20} or Minimum Required Length (MRL)~\cite{34,35,36, 37, 43}. This early classifier can classify an incoming time series as soon as MPL is available.         
\subsubsection{\textbf{Shapelet based early classification}} 
A family of early classification approaches~\cite{4, 10,11, 13, 14, 15, 17, 19, 24, 50, 51, 52} focused on obtaining a set of key shapelets from the training dataset and utilizing them as class discriminatory features of the time series. As there are many shapelets in the dataset, the different approaches attempted to select only those shapelets that can provide maximum earliness and uniquely manifest the class label. These selected shapelets are matched with the incoming time series, and the class label of best matched shapelet is assigned to the time series.      
\subsubsection{\textbf{Model based early classification}}
Another set of early classification approaches~\cite{7, 8, 25, 26, 30, 46, 47, 54} proposed mathematical models based on conditional probabilities. The approaches obtain these conditional probabilities by either fitting a discriminative classifier or using generative classifiers on training. Some of these early classification approaches have also developed a cost-based trigger function to make reliable predictions.   
         
\subsubsection{\textbf{Miscellaneous approaches}} 
The early classification approaches that do not qualify any of the above mentioned categories, are included here. Some of these approaches employed deep learning techniques~\cite{22,32}, reinforcement learning~\cite{31, 49}, and so on~\cite{3, 12, 16, 18}. 

\begin{table*}
\centering
\caption{Summary of the early classification approaches for time series. }
\begin{tabular}{| >{\centering\arraybackslash}m{1cm}|>{\centering\arraybackslash}m{4cm}|m{7.5cm}|c|>{\centering\arraybackslash}m{1.5cm}|}
\hline
\textbf{Paper} &  \textbf{Classifier used}     & \textbf{Datasets for experimental evaluation}                                                              & \textbf{\begin{tabular}[c]{@{}c@{}}Type of\\ time series\end{tabular}} & \textbf{Category}                                                                                   \\ \hline

\cite{1}                                                                                   & Decision tree and rule-based classifier & ECG~\cite{UCR},  synthetic control~\cite{UCI}, DNA sequence~\cite{UCI}                                                   & \multirow{5}{*}{UTS}                                                   & \multirow{16}{*}{\begin{tabular}[c]{@{}c@{}}Prefix \\ based \\ early \\ classification\end{tabular}} \\ \cline{1-3}

\cite{2}                                                                                             & 1-NN                         & 7 UCR datasets                                                                                             &                                                                        &                                                                                                     \\ \cline{1-3}

\cite{5}                                                                                       & 1-NN                         & 7 UCR datasets                                                                                             &                                                                        &                                                                                                     \\ \cline{1-3}

\cite{29}                                                                                      & Gaussian Process (GP) classifier                         & 45 UCR datasets                                                                                             &                                                                        &                                                                                                     \\ \cline{1-3}

\cite{36}                                                                                & GP classifier                         & River dataset~\cite{riverd}                                                                                             &                                                                        &                                                                                                     \\ \cline{1-4}

\cite{20}                                                                                         &        1-NN                      & Wafer and ECG~\cite{waferandecg}, character trajectories~\cite{UCI}, robot execution failures~\cite{UCI} & \multirow{9}{*}{MTS}                                                                   &     \\ \cline{1-3}

\cite{34}                                                                                      & GP classifier                         & Human activity classification (collected), NTU RGB+D~\cite{ntu}, daily and sports activities~\cite{UCI}, heterogeneity human activity recognition~\cite{UCI}  &   &             \\ \cline{1-3}                                        

\cite{35}                                                                                    & GP and Hidden Markov Model (HMM) classifiers                      & Road surface classification (collected),  PEMS-SF~\cite{UCI}, heterogeneity human activity recognition~\cite{UCI}, gas mixtures detection~\cite{UCI} &      & \\ \cline{1-3} 

\cite{43}                                                                                     & GP classifier                         & Hydraulic system monitoring~\cite{UCI},  PEMS-SF~\cite{UCI}, daily and sports activities~\cite{UCI} &      &
\\ \hline

\cite{4}                                                                          &          Closest shapelet using Euclidean Distance (ED)                  & 7 UCR datasets &    \multirow{6}{*}{UTS}  &    \multirow{20}{*}{\begin{tabular}[c]{@{}c@{}}Shapelet \\  based \\ early \\ classification\end{tabular}}  \\ \cline{1-3} 

\cite{10}                                                                              &   Closest shapelet using ED                         & 20 UCR datasets  &      &   \\ \cline{1-3}

\cite{50}                                                                              &   CNN                        & 12 UCR datasets  &      &   \\ \cline{1-3}

\cite{51}                                                                      &   Closest shapelet using ED                       & 35 UCR datasets  &      &   \\ \cline{1-3}

\cite{52}                                                                            &   Closest shapelet using Trend-based ED                       & 16 UCR datasets  &      &   \\ \cline{1-4}

\cite{11}                                                                              &   Closest multivariate shapelet using ED                         & 8 gene expression datasets~\cite{zaas2009gene, baranzini2004transcription} &   \multirow{12}{*}{MTS}   &   \\ \cline{1-3}

\cite{13}                                                                             &   Closest multivariate shapelet using ED                         &  Wafer and ECG~\cite{waferandecg}, 2 synthetic datasets  &      & \\ \cline{1-3}

\cite{14}                                                                             &   Closest multivariate shapelet using ED                         &  2 gene expression datasets~\cite{zaas2009gene}, ECG~\cite{goldberger2000physiobank} &      &  \\ \cline{1-3}

\cite{17}                                                                            &   Rule based and Query By Committee (QBC) classifiers                         &  Wafer and ECG~\cite{waferandecg}, 2 synthetic datasets &      &  \\ \cline{1-3}

\cite{19}                                                                         &  Decision tree                         &  Gene expression dataset~\cite{baranzini2004transcription}, Wafer and ECG~\cite{waferandecg}, robot execution failures~\cite{UCI} &      &  \\ \cline{1-3}

\cite{38}                                                                             &          Closest multivariate shapelet using ED               & Wafer and ECG~\cite{waferandecg}, Character trajectories~\cite{UCI}, Japanese vowels~\cite{UCI}, uWaveGestureLibrary~\cite{UCI}   &      &   \\ \cline{1-3}

\cite{24}                                                                            &  Decision tree and random forest                        &  ICU data of 2127 patients (collected) &      &   \\ \hline

\cite{7}                                                                      &  Quadratic Discriminant Analysis (QDA)                         & 1 synthetic and 4 UCR datasets  &   \multirow{13}{*}{UTS}  & \multirow{16}{*}{\begin{tabular}[c]{@{}c@{}}Model \\  based \\ early \\ classification\end{tabular}}    \\ \cline{1-3}

\cite{6}                                                                & Linear Support Vector Machines (SVM) and Local QDA                       &  15 UCR datasets &      &  \\ \cline{1-3}

\cite{25}                                                                 & Naive Bayes and Multi Layer Perceptron                      &  TwoLeadECG~\cite{UCR} &      &  \\ \cline{1-3}

\cite{26}                                                               & HMM and iHMM                      &  Japanese vowel speaker~\cite{UCI} &      &  \\ \cline{1-3}

\cite{47}                                                              & GP and SVM                      & 45 UCR datasets, IBEX35 stock &      &  \\ \cline{1-3}

\cite{44}                                                               & Linear SVM                      & CBF~\cite{UCR}, control charts~\cite{UCI}, character trajectories~\cite{UCI}, localization data for person activity~\cite{UCI} (after preprocessing)  &      &  \\ \cline{1-3}

\cite{28}                                                      & SVM                      & 76 UCR datasets &      &  \\ \cline{1-3}

\cite{30}                                                   & GP and SVM                      & 45 UCR datasets &      &  \\ \cline{1-3}

\cite{46}                                                  & GP                  & 45 UCR datasets &      &  \\ \cline{1-3}

\cite{54}                                                     &  GP, SVM, and Naive Bayes                 & 15 UCR datasets &      &  \\ \cline{1-3}

\cite{33}                                                     & Two-tier classifier using variants of SVM, DTW                  & 45 UCR datasets, PLAID~\cite{gao2014plaid}, ACS-F1~\cite{gisler2013appliance} &      &  \\ \cline{1-4}

\cite{8}                                                                & SVM                        &  Gas dataset (collected) &  MTS   &  \\ \hline

\cite{3}                                                                        &  Adaboost ensemble classifier                         & CBF~\cite{UCR}, control charts~\cite{UCI}, trace~\cite{UCR}, auslan~\cite{UCI}  &   \multirow{4}{*}{UTS}  & \multirow{10}{*}{\begin{tabular}[c]{@{}c@{}} Miscellaneous \\  approaches \end{tabular}}    \\ \cline{1-3}

\cite{31}                                                                          &  Reinforcement learning agent            & 3 UCR datasets  &     &     \\ \cline{1-3}

\cite{32}                                                                         &        Combination of CNN and LSTM      & 46 UCR datasets  &     &     \\ \cline{1-3}

\cite{53}                                                                           &       SVM and Neural network      & Bearing faults dataset   &     &     \\ \cline{1-4}

\cite{49}                                                                           & Deep Q-Network                         & Living organisms dataset
  &  \multirow{6}{*}{MTS}   &    \\ \cline{1-3}

\cite{18}                                                                        &   Hybrid model using HMM and SVM                         & 5 gene expression datasets~\cite{baranzini2004transcription}  &      &  \\ \cline{1-3}

\cite{16}                                                       & Stochastic process        &  Auslan~\cite{UCI}, PEMS-SF~\cite{UCI}, motion capture~\cite{mocap} &      &  \\ \cline{1-3}

\cite{21}                                                      & Stochastic process         &  Motion capture~\cite{mocap}, NTU RGB+D~\cite{ntu},  UT Kinect-Action~\cite{xia2012view} &      &  \\ \cline{1-3}

\cite{22}                                                                  &  Combination of CNN and LSTM                         &  Wafer and ECG~\cite{waferandecg}, auslan~\cite{UCI} &     &  
\\ \hline

\end{tabular}
\vspace{0.7cm}
\label{summary1}
\end{table*}

\vspace{0.3cm}
\section{Prefix based early classification}\label{pre}
This section discusses the prefix based early classification approaches in detail. 
The first notable prefix based approach was proposed in~\cite{1}. The authors in~\cite{1} introduced two novel methods, Sequential Rule Classification (SCR) and Generalize Sequential Decision Tree (GSDT), for early classification of symbolic sequences. For a given training dataset, SCR method first extracts a large number of sequential rules from the different lengths of prefix spaces and then selects some top-$k$ rules based on their support and prediction accuracy. These selected rules are used for the early classification.        

We split the prefix based approaches into two groups according to their MPL computation methods. In the first group, the approaches~\cite{2,5,20} developed a concept of Reverse Nearest Neighbor (RNN) to compute MPL of time series. In the second group of approaches~\cite{29, 34,35,36}, the authors employed a probabilistic classifier first to obtain posterior class probabilities and then utilized them for MPL computation.  

\vspace{0.3cm}
\subsection{MPL computation using RNN}
We first discuss the concept of RNN for the time series data and then describe the approaches that have been using RNN for MPL computation. Let $D$ is a labeled time series dataset with $N$ instances of length $T$. According to~\cite{2}, RNN of a time series $\mathbf{X} \in D$ is a set of those time series which have $\mathbf{X}$ in their nearest neighbors. It is mathematically given as 
\begin{align}\nonumber
 RNN^t (\mathbf{X}) = \{\mathbf{X}{'} \in D \big | \mathbf{X} \in NN^t(\mathbf{X}{'})\},
\end{align}
where, $t$ is the length of $\mathbf{X}$ in the prefix space and $t=T$ in full-length space. 

To compute MPL of the time series $\mathbf{X}$, the authors~\cite{2} compares $RNN$ of full-length space with $RNN$ of prefix spaces. The MPL of $\mathbf{X}$ is set to $t$ if the following conditions are satisfied: (1) $RNN^t(\mathbf{X})= RNN^T(\mathbf{X}) \neq \phi$, (2) $RNN^{t-1}(\mathbf{X}) \neq RNN^T(\mathbf{X}),$ and (3) $RNN^{t{'}}(\mathbf{X}) = RNN^T(\mathbf{X}),$
where $t \leq t{'} \leq T$. Further, if $RNN^T(\mathbf{X}) \neq \phi$ then MPL of $\mathbf{X}$ is equal to $T$. Above conditions check the stability of RNN using prefix of $\mathbf{X}$ with length $t$.

Xing \textit{et al.}~\cite{2} developed two different algorithms, Early 1-NN and Early Classification of Time Series (ECTS), for UTS data. Early 1-NN algorithm computes the MPLs for all the time series of the training dataset. These computed MPLs are first arranged in ascending by their lengths and then used for early classification of the testing time series $\mathbf{X}{'}$. Let $m$ be a least value of the computed MPLs. As soon as the number of data points in $\mathbf{X}{'}$ becomes equal to $m$, Early 1-NN starts its classification. It first computes 1-NN of $\mathbf{X}{'}$ with $m$ data points as follows
\begin{align}
 NN^m(\mathbf{X}{'}) = \underset{\mathbf{X} \in D_{mpl}}{\text{argmin}}\{dist(\mathbf{X}{'}[1,m], \mathbf{X}[1,m])\},
\end{align}
where, $D_{mpl}$ is the dataset of those time series whose MPLs are at most $m$. The function $dist(\cdot)$ computes the euclidean distance between the time series. 

Early 1-NN has two major drawbacks: i) each time series can have different MPL, and ii) computed MPLs are short and not robust enough due to the  overfitting problem of 1-NN. To overcome these drawbacks, ECTS algorithm~\cite{2} first clusters the time series based on their similarities in full-length space and then computes only one MPL for each cluster.
In~\cite{5}, the authors presented an extension of ECTS, called as \textit{Relaxed} ECTS, to find shorter MPLs. It relaxed the stability condition of RNN while computing MPLs for the clusters. 

In~\cite{20}, the authors proposed an MTS Early Classification based on PAA (MTSECP) approach where PAA stands for Piecewise Aggregated Approximation~\cite{keogh2001dimensionality}. MTSECP first applies a center sequence method~\cite{li2015piecewise} to transform each MTS instance of the dataset into UTS and then reduces the length of the transformed UTS by using PAA method. 

\subsection{MPL computation using posterior probabilities}
\begin{figure}[h]
\centering
\includegraphics[scale=1]{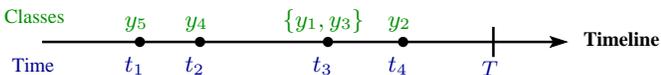}
\caption{Illustration of the discriminative MPLs for five different class labels, \textit{i.e.,} $y_1, y_2, \cdots, y_5$. }
\label{mpl}
\vspace{0.3cm}
\end{figure}

Apart from RNN, some researchers have also utilized the posterior probabilities for MPL computation of time series. This group of early classification approaches computes a class discriminative MPL for each class label of the dataset. For a given training dataset, these approaches fit a probabilistic classifier in the prefix space of length $t$, where $1 \leq t \leq T$. The probabilistic classifier provides posterior class probabilities for each time series of the training dataset. 
The class discriminative MPL for the class label $y$ is set to $t$ if 
\begin{align}
 \mathbf{A}^t_y \geq \alpha \cdot \mathbf{A}^T_y,
\end{align}
where, $\mathbf{A}^t_y$ and $\mathbf{A}^T_y$ are the training accuracy for class label $y$ in the prefix space of length $t$ and full-length space, respectively. The parameter $\alpha$ denotes a desired level of accuracy of the early classification and $0 < \alpha \leq 1$. 
Fig.~\ref{mpl} shows an example of the discriminative MPLs for five different classes along the progress of time series. 

Mori \textit{et al.}~\cite{29} proposed an Early Classification framework based on DIscriminativeness and REliability (ECDIRE) of the classes over time. ECDIRE employed GP classifier~\cite{GP} to compute the class discriminative MPLs. It also computes a threshold for each class label to ensure the reliability of the predictions. 
The threshold for the class label $y$ is computed as
\begin{align}
 \theta_{t,y} = \underset{\mathbf{X}\in D_y}{\text{min}} \{p_1^t(\mathbf{X})-p_2^t(\mathbf{X})\},
\end{align}
where, $p_1^t(\mathbf{X})$ and $p_2^t(\mathbf{X})$ denote first and second highest posterior probabilities for a training time series $\mathbf{X}$ using the prefix of length $t$, respectively. The dataset $D_y$ consists the time series that are correctly classified in class $y$. 

In~\cite{36}, the authors developed a game theoretic approach for early classification of Indian rivers using the time series of water quality parameters such as pH value, turbidity, dissolved oxygen, \textit{etc.} They formulated an optimization that helps to compute the class-wise MPLs while maintaining $\alpha$ accuracy. 

The authors in~\cite{35,37, 43} attempted to classify an incoming MTS as early as possible with at least $\alpha$ accuracy. They focused handling a special type of MTS collected by the sensors of different sampling rate. The proposed approaches~\cite{35,37, 43} first estimate the class-wise MPLs for each component (\textit{i.e.,} time series) separately and then develop a class forwarding method to early classify an incoming MTS using the computed MPLs. On the other hand, the approach~\cite{43} proposed a divide-and-conquer based method to handle the different sampling rate components.

\begin{table*}[h]
\caption{Analysis of prefix based early classification approaches using their strength, limitation, and major concern.}
\centering
\begin{tabular}{|>{\centering\arraybackslash}m{1cm}|>{\centering\arraybackslash}m{4.5cm}|>{\centering\arraybackslash}m{5.8cm}|>{\centering\arraybackslash}m{4.7cm}|}
\hline
\textbf{Paper} & \textbf{Strength}                                                  & \textbf{Limitation}         & \textbf{Major Concern}                                                          \\ \hline
\cite{1} & Robust entropy-based utility measure & Time series has to be discretized properly & Symbolic representation of time series  \\ \hline
\cite{2} & Simple model construction with reliability assurance & Cluster separation can not be guaranteed for small datasets &  MPLs computation using hierachical clustering \\ \hline        
\cite{5}  & Simple and effective model with 1-NN & Overfitting problem for small datasets & Relaxation of RNN stability condition \\ \hline
\cite{29} & Class discriminative MPLs with reliability threshold & No check for stability while learning MPLs &  Avoiding unnecessary predictions before the availability of sufficient data \\ \hline 
\cite{36} & Game theory based tradeoff optimization & Need to set several parameters which requires domain knowledge & Computation of class-wise MPLs using probabilistic classifier \\ \hline
\cite{20} & Abililty to handle MTS using center sequence transformation & Approximation of segments causes to lose identifiable information of the classes & Transformation of MTS into UTS \\ \hline
\cite{34} & Robust to faulty components of MTS & Number of samples in the faulty components should be equal to non-faulty & Selection of relevant components of MTS \\ \hline
\cite{35} & Capable to classify MTS with varying length of components & Component of highest sampling rate is required to have full length before starting the prediction & Designing of the class forwarding method to incorporate correlation \\ \hline
\cite{43} & Ability to handle MTS generated from the sensors of different sampling rate &  No check for stability while learning MPLs & Utilization of correlation  \\ \hline
\end{tabular}
\label{prefixbased}
\vspace{0.3cm}
\end{table*}

Gupta \textit{et al.}~\cite{34} extended the concept of early classification for the MTS with faulty or unreliable components. They proposed a Fault-tolerant Early Classification of MTS (FECM) approach to classify an ongoing human activity by using the MTS of unreliable sensors. FECM first identifies the faulty components using Auto Regressive Integrated Moving Average (ARIMA) model~\cite{box2015time} whose parameters are learned from the training instances. A utility function is also developed to optimize the tradeoff between accuracy $\mathbf{A}_t$ and earliness $\mathbf{E}$, as given below
\begin{align}
 \mathcal{U}(\mathbf{X}[1,t]) = \frac{2 \times \mathbf{A}_t \times \mathbf{E} }{\mathbf{A}_t + \mathbf{E}}.
\end{align}
The accuracy $\mathbf{A}_t$ is computed using the confusion matrix obtained by applying $k$-means clustering on the training dataset. Next, the MPL of a time series $\mathbf{X}$ is computed as
\begin{align}
\text{MPL} (\mathbf{X}) = \underset{1 \leq  t \leq T }{\text{argmax}} \{\mathcal{U}(\mathbf{X}[1,t])\}.
\end{align}
Finally, FECM employed the kernel density estimation method for learning the class-wise MPLs.    

\subsection{Critical analysis}
The prefix based early classification approaches are simple and easy to understand. However, due to a lack of interpretability in the early classification results, these approaches are not suitable for medical applications. As we already categorized the approaches based on their similarities, we now review them based on the following parameters: strength, limitation, and major concern. These parameters are sufficient to make a critical or insightful analysis of an early classification approach. Table~\ref{prefixbased} presents the analysis of prefix based approaches using the above mentioned parameters.   

\vspace{0.3cm}
\section{Shapelet based Early classification}\label{sha}
This section presents a detailed review of the approaches that have used shapelets for the early classification of time series. The authors in~\cite{ye2009time, ye2011time} have successfully implemented the idea of shapelets for time series classification, which became the motivation point for many researchers to utilize the shapelets for achieving the earliness in the classification. 

For a given training dataset, the early classification approaches first extract all possible subsequences (segments) of the time series with different lengths and then evaluate their quality and earliness to obtain a set of key shapelets. 
Let $S=\{s, l, \delta, y\}$ be a shapelet as discussed in the fundamentals. The distance threshold $\delta$ of the shapelet $S$ is computed as

\begin{align}\label{mindist}
 d = \underset{s{'} \sqsubseteq \mathbf{X}, |s{'}|=|s|}{\text{min}} \big \{dist(s, s')\big \},
\end{align}
where, the symbol $\sqsubseteq$ is used to select a subsequence from the set of all subsequences of $\mathbf{X}$. The authors in the existing approaches~\cite{10, 11, 13, 14, 15, 17} have developed different methods for computing the distance threshold $\delta$. The shapelets are filtered out based on their utility to obtain the key shapelets, which are later used to early classify an incomplete time series. 

\begin{figure}[h]
\centering
\includegraphics[scale=1]{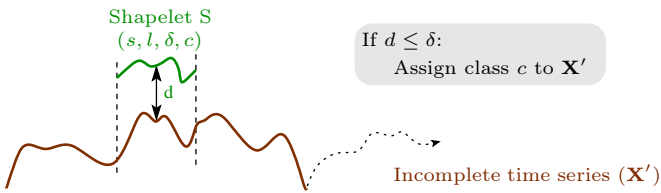}
\caption{Early classification of the time series using shapelet. }
\label{shape}
\end{figure}

An example of the early classification using shapelets is illustrated in Fig.~\ref{shape}. The class label of the shapelet $S$ is assigned to $\mathbf{X}{'}$ if the distance $d$ between $\mathbf{X}{'}$ and $S$ is less than its pre-computed threshold $\delta$. The shapelet based approaches are divided into two groups according to the key shapelet selection methods. 

\subsection{Key shapelet selection using utility measure}
The first work to address the early classification problem using shapelets is presented in~\cite{4}. The authors developed an approach called Early Distinctive Shapelet Classification (EDSC), which utilizes the local distinctive subsequences as shapelets for the early classification. EDSC consists of two major steps: \textit{feature extraction} and \textit{feature selection}. In former step, it first finds all local distinctive subsequences from the training dataset and then computes a distance threshold $\delta$ for each subsequence. 
Next, in feature selection step, the authors select key shapelets based on their utility. In EDSC, the utility of the shapelet $S$ is computed using its precision $P$ and weighted recall $R_w$, as given below 
\begin{align}\nonumber
 U(S) = \frac{2 \times P(S) \times R_w(S)}{ P(S) + R_w(S)}.
\end{align}
The precision $P(S)$ captures the class distinctive ability of the shapelet on the training dataset. On the other, the weighed recall $R_w(S)$ captures earliness and frequency of shapelets in the training instances. 

Ghalwash \textit{et al.}~\cite{10} presented an extension of EDSC called as Modified EDSC with Uncertainty (MEDSC-U) estimate. The uncertainty estimate indicates the confidence level with which the prediction decision is made, and if it is less than some user-defined confidence level then the decision may be delayed even after a shapelet is matched. The work in~\cite{52} also introduced an Improved version of EDSC (IEDSC) with a trend-based euclidean distance. 

In~\cite{11}, the authors utilized shapelets for early classification of gene expression data. A Multivariate Shapelets Detection (MSD) method is proposed to classify an incoming MTS by extracting the key shapelets from the training dataset. MSD finds several multivariate shapelets from all dimensions of MTS with same start and end points.  

Lin \textit{et al.}~\cite{19} developed a Reliable EArly ClassifiTion (REACT) approach for MTS where some of the components are categorical along with numerical. REACT first discretizes the categorical time series and then generates their shapelets. It employed a concept of Equivalence Classes Mining~\cite{lo2008mining} to avoid redundant shapelets. Finally, a change-point based distance measure is proposed in~\cite{51}, to compute similarity between the time series and shapelet. Besides that, deep learning techniques~\cite{50} are also employed for extracting the multi-scale shapelets based on a cost function.
 
\begin{table*}[h]
\caption{Analysis of shapelet based early classification approaches using their strength, limitation, and major concern.}
\centering
\begin{tabular}{|>{\centering\arraybackslash}m{1cm}|>{\centering\arraybackslash}m{5cm}|>{\centering\arraybackslash}m{4.5cm}|>{\centering\arraybackslash}m{5.5cm}|}
\hline
\textbf{Paper} & \textbf{Strength}                                                  & \textbf{Limitation}         & \textbf{Major Concern}                                                          \\ \hline
\cite{4} & Highly interpretable shapelets  & No assurance of classification accuracy during training  & Utility based shapelet selection \\ \hline
\cite{10} & Uncertainty estimates with highly interpretable shapelets  & Generate huge number of candidate shapelets of varying length
 & Computation of shapelet rank by incorporating accuracy and earliness \\ \hline
 \cite{50} & Multi-scale deep features (shapelets) of time series & Need to set several parameters at each layer of the model 
 & Automatic feature extraction using deep learning models \\ \hline
 \cite{51} & Robust to distance information noise while extracting shapelets & Extracted shapelets tend to lose natural interpretability
 &  Computation of distance between shapelet and time series in change-point space \\ \hline
 \cite{52} & Diverse and highly interpretable shapelets & Computationally inefficient due to large number of shapelets  & Developing trend based euclidean distance to incorporate diversity \\ \hline
 \cite{11} & Multivariate shapelets with high utility  & Extracted shapelets can not have variable length of dimensions  & Separate distance threshold along each dimension of MTS \\ \hline
 \cite{13} & Ability to handle imbalance distribution of the instances among classes & Limited to binary classification  & Clustering based core shapelet selection  \\ \hline 
 \cite{14} & Ability to identify relevant dimensions of MTS & Data points of time series must be obtained at regular interval
&  Formulation of convex optimization problem for key shapelet selection \\ \hline
\cite{17} & Utilization of internal relationship among the dimensions & Quality of shapelets heavily depends on the employed clustering method
 & Evaluation strategy to check the quality of shapelets \\ \hline
 \cite{19} & Capable enough to classify a time series with categorical samples  & Computationally inefficient  & Pattern discovery using sequential and simultaneous combinations of shapelets \\ \hline
 \cite{38} & Highly interpretable shapelets with confidence estimates about early prediction & Limited to binary classification & Key-point based shapelet extraction \\ \hline
 \cite{24} & Ability to classify asynchronous MTS  & No assurance of reliability and need to set several parameters & Inclusion of short-term trend in features
\\ \hline
\end{tabular}
\label{shapeletbased}
\vspace{0.3cm}
\end{table*}
\subsection{Key shapelet selection using clustering}
He \textit{et al.}~\cite{13} attempted to solve an imbalanced class problem of ECG classification where training instances in the abnormal class are much lesser than normal. The authors addressed this problem in the framework of early classification and proposed a solution approach, called as Early Prediction on Imbalanced MTS (EPIMTS). 
At first, the candidate shapelets are clustered using Silhouette Index method~\cite{rousseeuw1987silhouettes}. Later, the shapelets in the clusters are ranked according to a Generalized Extended F-Measure (GEFM). The shapelet with maximum rank is used to present the respective cluster. For a shapelet $S$, GEFM is computed as
\begin{align}\label{gefm}
GEFM (S) = \frac{1}{w_0/\mathbf{E}(S) + w_1/P(S) + w_2/R(S)}, 
\end{align}
where, the weight parameters $w_0, w_1,$ and $w_2$ are used to control the importance of earliness $\mathbf{E}$, precision $P$, and recall $R$, respectively. 

In~\cite{14}, the authors proposed an approach, called as Interpretable Patterns for Early Diagnosis (IPED), for studying the viral infection in humans by using their gene expression data. Similar to MSD, IPED also extracts multivariate candidate shapelets from the training MTS, but it allows to have a multivariate shapelet with different start and end points in the dimensions. IPED computes an information gain based distance threshold for the shapelets. 

One of the major drawback of MSD~\cite{11}, EPIMTS~\cite{13}, and IPED~\cite{14} approaches, is the avoidance of the correlation that exists among the shapelets of different components of MTS. Such a correlation helps to improve the interpretability of the shapelets. To overcome this drawback, the authors in~\cite{15, 17} developed an approach, called as Mining Core Features for Early Classification (MCFEC), where core features are the key shapelets. 
  
The authors in~\cite{38} presented a Confident Early Classification framework for MTS with interpretable Rules (CECMR), where the key shapelets are extracted by using a concept of the local extremum and turning points. Local extemum point of a time series $\mathbf{X}$ is $\mathbf{X}[t]$ if 
\begin{align}\nonumber
 \mathbf{X}[t] > \mathbf{X}[t-1] \ &\& \ \mathbf{X}[t] > \mathbf{X}[t+1] \\ \nonumber
&\text{or} \\ \nonumber
  \mathbf{X}[t] < \mathbf{X}[t-1] \ &\& \ \mathbf{X}[t] < \mathbf{X}[t+1]. \nonumber 
\end{align}
where, $1 \leq t \leq T$. Next, the turning point of $\mathbf{X}$ is $\mathbf{X}[t]$ if the following condition holds.
\begin{align}\nonumber
 (\mathbf{X}[t+1] - 2 \mathbf{X}[t] + \mathbf{X}[t-1]) > \sum_{i=1}^{T} \frac{\mathbf{X}[i] - \mathbf{X}[i-1]}{T-1}.
\end{align}
CECMR first discovers interpretable rules from the sets of candidate shapelets and then estimates the confidence of each rule to select the key shapelets. 
Recently, the shapelets are also adopted for estimating an appropriate time to transfer a patient into ICU by using the MTS of physiological signs~\cite{24}.

\subsection{Critical analysis}
The shapelets help to improve interpretability of the classification results~\cite{4, 10, 14}, which enhances the adaptability of an early classification approach for the real-world applications such as health informatics and industrial process monitoring. In shapelet based approaches, the authors focused on extracting a set of perfect or key shapelets from the given training dataset. Ideally, a perfect shapelet should be powerful enough to distinguish all the time series of one class from that of other classes. The researchers therefore put their efforts towards developing a proper criterion that can provide a set of effective shapelets (if not perfect)~\cite{13, 14, 19,24}. 

The clustering of candidate shapelets has been proven a good idea to select more distinctive key shapelets than those selected by the utility measure. However, its credit goes to GEFM, which ranks the shapelets based on their distinctiveness, earliness, and frequency. The approaches~\cite{17, 15, 38} utilize the correlation among the components of the shapelets, which improved their earliness and interpretability to a great extent. Further, a detailed analysis of the shapelet based approaches is presented in Table~\ref{shapeletbased}.
 
\section{Model based early classification}\label{mod}
This section discusses the model based early classification approaches for time series data. Unlike prefix and shapelet based approaches, the model based approaches~\cite{6, 7, 27, 30} formulate a mathematical model to optimize the tradeoff between earliness and reliability. 
We divide these approaches into two following groups based on the type of adopted classifier.

\subsection{Using discriminative classifier}
In~\cite{8}, the authors developed an ensemble model to recognize the type of gas using an incomplete 8-dimensional time series generated from a sensors-based nose. The ensemble model consists of a set of probabilistic classifiers with a reject option that allows them to express their doubt about the reliability of the predicted class label. The probabilistic classifier assigns a class label $y{'}$ to the incomplete time series $\mathbf{X}{'}$ as given below

\begin{align}
 y{'} = \underset{y \in \{y_1, y_2, \cdots, y_k\}}{\text{argmax}} \{p(y|\mathbf{X}{'})\}.
\end{align}

\noindent If $p(y{'}|\mathbf{X}{'})$ is close to $0.5$ then the classifier chooses reject option to express the doubt on the class label $y{'}$. 
Another work in~\cite{44} focused on minimizing response time to obtain the earliness in the classification. This work developed an empirical function that optimizes the earliness with a high degree of confidence.

\begin{table*}[t]
\caption{Analysis of model based early classification approaches using their strength, limitation, and major concern.}
\centering
\begin{tabular}{|>{\centering\arraybackslash}m{1cm}|>{\centering\arraybackslash}m{5cm}|>{\centering\arraybackslash}m{5cm}|>{\centering\arraybackslash}m{5cm}|}
\hline
\textbf{Paper} & \textbf{Strength}                                                  & \textbf{Limitation}         & \textbf{Major Concern}    \\ \hline                                                      
\cite{7} & Guarantee of the desired level of accuracy with earliness & Several iterations are required for setting an appropriate value of reliability threshold
& Construction of the optimal decision rules \\ \hline
\cite{25} & Capable to estimate future time step where correct prediction is expected  & Cost function heavily relies on the clustering accuracy
 & Designing of the cost-based trigger function \\ \hline 
 \cite{26} & Ability to work on streaming data  & High domain dependency & Likelihood based similarity measure between the time series \\ \hline
 \cite{47} & Simple and effective model & Non-convex cost function & Stopping rules using selected posterior probabilities \\ \hline
 \cite{44} & Ensemble classifier with reject option & Computationally inefficient & Formulation of the convex optimization problem for training  \\ \hline
 \cite{28} &  Adaptive to the testing time series &  Impose unnecessary computations due to non-myopic property & Designing of the trigger function \\ \hline
 \cite{30} & Highly relevant cost functions &  Non-convex cost function & Stopping rules using all posterior probabilities \\ \hline
 \cite{46} & Effective model with the adaptive learning of important parameters & Single confidence threshold is not sufficient for generalization
& Fusion of multiple classifiers for making the early decision \\ \hline
\cite{54} & Class-wise safeguard points with the assurance of desirable accuracy & No check for stability while discovering safeguard points
& Avoidance of premature predictions \\ \hline
\cite{33} & Robust to varying start time of events  & Difficult to find a suitable interval length to capture identifiable patterns & Master-slave architecture for finding an optimal decision time \\ \hline
\cite{8} & Stable ensemble classifier & Limited to work with probabilistic classifiers only & Providing a reject option to express the doubt about reliability of prediction
\\ \hline
\end{tabular}
\label{modelbased}
\vspace{0.3cm}
\end{table*}

Dachraoui \textit{et al.}~\cite{25} proposed a non-myopic early classification approach where the term \textit{non-myopic} means, at each time step $t$, the classifier estimates an optimal time $\tau^*$ to give an assurance of the reliable prediction in the future. For an incomplete time series $\mathbf{X}_t^{'}$ with $t$ data points, the optimal time $\tau^{*}$ is calculated using following expression
\begin{align}\nonumber
 \tau^{*} = \underset{\tau \in \{0, 1,\cdots, T-t\}}{\text{argmax}} f_\tau(\mathbf{X}^{'}_t),
\end{align}
where, the function $f_\tau(\mathbf{X}^{'}_t)$ estimates an expected cost of the future time step $t+\tau$. The authors in~\cite{28} pointed out two weaknesses of the work~\cite{25}: i) assumption of low intra-cluster variability, and ii) clustering with the complete time series. In~\cite{28}, two different algorithms (\textit{NoCluster} and \textit{2Step}) are introduced to overcome these weaknesses while preserving the adaptive and non-myopic properties of the classifier. 

Mori \textit{et al.}~\cite{27} proposed an EarlyOpt framework that formulates a stopping rule using the two highest posterior probabilities obtained from the classifiers. The main objective of EarlyOpt is to minimize the cost of prediction by satisfying the stopping rule. 
The authors also conducted a case study for IBEX $35$ stock market in~\cite{47}.
In another work~\cite{30}, they developed different stopping rules by using the class-wise posterior probabilities. Besides that, the authors in~\cite{46, 54} computed a confidence-based threshold to indicate the data sufficiency for making the early prediction.  

The authors in~\cite{33} introduced a two-tier early classification approach based on the master-slave paradigm. In the first tier, the slave classifier computes posterior probabilities for each class label of the dataset and constructs a feature vector for each training time series. Let $p_{1:k}(\mathbf{X})$ be a set of $k$ posterior probabilities which is obtained for the training time series $\mathbf{X}$. The feature vector for $\mathbf{X}$ is given as

\begin{align}
  F_\mathbf{X} = \{y(\mathbf{X}), p_{1:k}(\mathbf{X}), \Delta_\mathbf{X}\}, 
\end{align}

\noindent where, $y(\mathbf{X})$ is the most probable class label and $\Delta_\mathbf{X}$ is the difference between first and second highest posterior probabilities. The feature vector is passed to the master classifier. In the second tier, the master classifier checks the reliability of the probable class label and makes the decision.   
 
\subsection{Using generative classifier}
The authors in~\cite{6} formulated a decision rule to classify an incomplete time series with some predefined reliability threshold. 
Two generative classifiers linear SVM and QDA~\cite{srivastava2007bayesian} with the formulated decision rule, are adopted to provide the desired level of accuracy in the early classification. 
An Early QDA model is proposed in~\cite{7} assuming that the training instances have Gaussian distribution. This assumption helps estimate parameters (\textit{i.e.,} mean and covariance) easily from training data. 

Antonucci \textit{et al.}~\cite{26} developed a generative model based approach for early recognition of Japanese vowel speakers using their speech time series data. The proposed approach employed an imprecise HMM (iHMM)~\cite{antonucci2015robust} to compute likelihood of intervals of incoming time series with respect to the training instances. 
For the reliable prediction, a class label is assigned to the time series only if the ratio of two highest likelihoods is greater than a predefined threshold.

\subsection{Critical analysis}
We found two exciting approaches~\cite{25,28} addressing the early classification problem with the non-myopic property. However, the computational complexity of these approaches is very high during classification. We analyzed that the generative classifier based early classification approaches are more complicated than those of discriminative classifiers. This work also analyzed the model based approaches by their strength, limitation, and major concern, as summarized in Table~\ref{modelbased}.

\section{Miscellaneous approaches}\label{mis}
This section covers the early classification approaches that do not meet the inclusion criteria of other categories. We split the included approaches into the following two groups: with and without tradeoff. 

\subsection{With tradeoff}
The authors in~\cite{31, 49} introduced a reinforcement learning based early classification framework using a Deep Q-Network (DQN)~\cite{mnih2015human} agent. The framework uses a reward function to keep balance between the accuracy and earliness. The DQN agent learns an optimal decision-making strategy during training, which helps pick a suitable action after receiving an observation in an incoming time series during the testing.    

In another work~\cite{32}, the authors developed a deep neural network based early classification framework that focused on optimizing the tradeoff by estimating the stopping decision probabilities at all time stamps of the time series. They formulated a new loss function to compute the loss of the classifier when a class label $y$ is predicted for an incomplete time series $\mathbf{X}^{'}_t$ using first $t$ data points. The loss at time $t$ is computed as

\begin{align}
 \mathcal{L}_t(\mathbf{X}^{'}_t, y; \beta) = \beta \mathcal{L}_a(\mathbf{X}^{'}_t, y) + (1 - \beta) \mathcal{L}_e(t),
\end{align}
where, $\beta$ is a tradeoff parameter to control the weights of the accuracy loss $\mathcal{L}_a(\cdot)$ and earliness loss $\mathcal{L}_e(\cdot)$. 

\subsection{Without tradeoff}
The first work that mentioned the early classification of time series is presented in~\cite{3}. The authors aimed to classify the incomplete time series, but they did not attempt to optimize the tradeoff between accuracy and earliness. In~\cite{3}, the time series is represented by its states such as increase, decrease, stay, over the time, and so on. The authors first divided the time series into segments such that the segment can capture only one particular state. Later, each segment is replaced by a predicate indicating the presence of the state in terms of \textit{True} or \textit{False}. Finally, the availability of the predicates in the time series is used for the early classification.

In~\cite{12, 18}, hybrid early classification models are presented by combining a generative model with a discriminative model. At first, HMM classifiers are trained over short segments (shapelets) of the time series to learn the distribution of patterns in the training data. Next, the trained classifiers generate an array of log likelihoods for the disjoint shapelets of the time series, which is passed as feature vector for training the discriminative classifiers. 

The work in ~\cite{16, 21} employed a stochastic process, called as Point Process model, to capture the \textit{temporal dynamics} of different components of MTS. 
At first, the \textit{temporal dynamics} of each component is extracted independently and then the \textit{sequential cue} that occurs among the components is computed to capture the temporal order of events. 

Recently, Huang \textit{et al.}~\cite{22} proposed a Multi-Domain Deep Neural Network (MDDNN) based early classification framework for MTS. MDDNN employed two widely used deep learning techniques including CNN and LSTM. It first truncates the training MTS up to a fixed time step and then gives it as input to a CNN layer which is followed by another CNN and LSTM layers. Frequency domain features are also calculated from the truncated MTS. Another work in~\cite{53}, computed a vector of statistical features from the incomplete time series for its classification.

 \begin{table*}[h]
\caption{Analysis of miscellaneous early classification approaches based on their strength, limitation, and major concern.}
\centering
\begin{tabular}{|>{\centering\arraybackslash}m{1cm}|>{\centering\arraybackslash}m{4.5cm}|>{\centering\arraybackslash}m{5.5cm}|>{\centering\arraybackslash}m{5cm}|}
\hline
\textbf{Paper} & \textbf{Strength}                                                  & \textbf{Limitation}         & \textbf{Major Concern}    \\ \hline                                                      
\cite{3} & Simple and effective model & No guarantee about the earliness  & Improvement of the classification accuracy \\ \hline
\cite{31} & Ability to classify UTS, MTS and symbolic sequence  & Difficult to understand the early classification problem as reinforcement learning
 & Designing reward function to make balance between accuracy and earliness \\ \hline
\cite{32} & Scalability to large dataset & Difficult to select appropriate values for the parameters  & Designing stopping rule using deep learning based posterior probabilities\\ \hline
\cite{53} & Computationally efficient model  & Statistical features lose the temporal information of the time series & Computing relevant features from the time series data \\ \hline
\cite{49} & Support to online learning & Complex formulation with large number of parameters & Designing reliable reward function \\ \hline
\cite{18} & Highly accurate model for the gene expression MTS & Computationally inefficient due to large number of segments & Combining generative and discriminative models \\ \hline
\cite{21} & Ability to capture identifiable patterns of an ongoing time series & No assurance of the reliable prediction & Utilization of the correlation \\ \hline
\cite{22} & Capable to extract features automatically & Unable to utilize correlation among the components of MTS   & Combining deep learning techniques for the early prediction of class label \\ \hline
\end{tabular}
\label{miscellaneous}
\vspace{0.3cm}
\end{table*}

Furthermore, the authors in~\cite{55, 56, 57} worked on video data for early recognition of an ongoing activity by extracting the time series features. In particular, the work in~\cite{55} first represented the human activity as histograms of spatio-temporal features obtained from the sequence of video frames. Later, these histograms are used to classify an ongoing activity.
The authors in~\cite{56} encoded the video frames into the histograms of oriented velocity, which are later used as features for early classification of the human activity. Hoai \textit{et al.}~\cite{57} introduced a maximum-margin framework for early detection of facial expressions by using the time series data of the partially executed events.

\subsection{Critical analysis}
A primary objective of the early classification approach is to build a classifier that can provide earliness while maintaining a desired level of reliability or accuracy. However, some approaches~\cite{12, 18, 16, 21, 3,22} do not ensure the reliability, but they are capable enough to classify an incomplete time series. Recently, the researchers in approaches~\cite{31, 49, 32, 22} have successfully employed reinforcement learning and deep learning techniques for the early classification. These approaches have unfolded a new direction for further research in this area. 
Analysis of the miscellaneous early classification approaches is presented in Table~\ref{miscellaneous}.     

\section{Discussion}\label{dis}
With the presented categorization of early classification approaches, one can get a quick understanding of the notable contributions that have been made over the years. After reviewing the literature, we found that most of the early classification approaches have appeared after ECTS~\cite{2}. Although the authors in~\cite{3} have attempted to achieve the earliness far before than ECTS, they did not maintain the reliability of class prediction. We therefore discussed such approaches in the without tradeoff category. Next, we discuss the challenges and recommendations of the reviewed studies.
\subsection{Challenges} 
After reviewing the included studies, we observed that the researchers have encountered four major challenges while developing the early classification approaches for time series data. These challenges are discussed below:    

\subsubsection{\textbf{Tradeoff optimization}}
 A most critical challenge before an early classifier is to optimize the tradeoff between accuracy and earliness. The studies in~\cite{34, 35, 36, 37, 43, 29} have attempted to address this challenge by learning a minimum required length of time series while maintaining a desired level of accuracy during the training. The researchers in~\cite{30, 27, 47} introduced stopping rules to decide the right time for early classification of an incomplete time series. Such a decision is evaluated through a cost function that ensures the balance between accuracy and earliness. A game theory model is adopted in~\cite{37} for tradeoff optimization. Recently, the deep learning techniques and reinforcement framework are also employed for optimizing the tradeoff~\cite{32, 31, 49}.              
 \subsubsection{\textbf{Interpretability}} In order to improve the adaptability of an early classification approach, the results should be interpretable enough to convince the domain experts, and thus interpretability becomes a challenge for the researchers. The studies in~\cite{4, 10, 50, 52, 11, 38, 13, 14} have primarily focused on extracting the interpretable features (shapelets) from the time series. In particular, the authors in~\cite{4, 10} selected the shapelets (subsequences) based on their utility and interpretability. The work in~\cite{52} developed a trend-based distance measure to find a diverse set of interpretable shapelets. 
  In~\cite{24}, the authors extracted the shapelets with short-term trends to improve the interpretability of results while classifying the irregular time series of physiological measurements. 
  \subsubsection{\textbf{Reliability}} It is a confidence measure that guarantees a certain level of accuracy in the early classification. Without the assurance of reliability, the early classification algorithm can not be used in real-world applications such as medical diagnostic and industrial process monitoring. In studies~\cite{7, 6, 49, 10, 11,  29, 51, 46, 54}, the authors have attempted to improve the reliability of early prediction. The authors in~\cite{51} developed a confidence area based criterion to ensure the reliability of early prediction. The approaches in~\cite{10, 11} learned a confidence threshold for each of the extracted key shapelet. The learned threshold is used to indicate the uncertainty estimate (reliability) of the shapelet for achieving the earliness. 
  The prefix based approaches in~\cite{1, 2, 5, 34, 35, 29} also ensure the reliability while learning the minimum prediction length, but they did not mentioned it explicitly.         
  \subsubsection{\textbf{Correlation}} There exists an inherent correlation among the components of MTS, which can help find the identifiable patterns at early stage. However, it is challenging to incorporate such correlation in the classification. The studies in~\cite{16, 21} employed point process models to early classify the MTS by using its temporal dynamics and sequence cue patterns. These patterns are capable enough to capture the correlation among components of MTS. In~\cite{35, 37, 43}, the researchers have introduced class forwarding methods to incorporate the correlation during early classification.

\subsection{Recommendations for future work} 
We briefly discuss some of the recommendations that have been provided by the reviewed studies. These recommendations help to conduct future research in this area to mitigate the limitations of the existing literature.   
 \subsubsection{\textbf{Prefix based approaches}} The studies in~\cite{1,2,5} suggested to employ probabilistic classifiers for obtaining shorter and effective MPLs. They also recommended to explore early classification methods for steaming time series with multiple class labels. A feature subset selection can be incorporated to improve the reliability of MPLs~\cite{20}. The work in~\cite{29} suggested to use an informative uncertainty measure in the learning phase for accommodating additional knowledge of the classes. The desired level of accuracy can be determined automatically by the knowledge of application domain~\cite{43}. The study in~\cite{34} recommended to extract relevant features from the MTS for its early classification.
 \subsubsection{\textbf{Shapelet based approaches}} Neural network models, especially LSTM, can deal with time series data in more natural way~\cite{50}. The studies in~\cite{4, 52} suggested to incorporate better similarity measures for improving the effectiveness of feature selection step. Fourier and wavelet transform techniques can be employed to obtain an useful combination of the features~\cite{19}. The authors in~\cite{11} indicated to incorporate a concept of maximal closed shapelets for pruning the redundant and smaller shapelets. The study in~\cite{17} suggested to utilize the sequential relationship between the components of MTS shapelets for achieving better earliness.
\subsubsection{\textbf{Model based approaches}} The study in~\cite{25} suggested to use the training instances for predicting the future decision cost without using any clustering method. The work in~\cite{47} recommended the use of genetic algorithms to automatically learn the shape of the stopping rule from the training data. The early classification can also be formulated as multi-objective optimization problem without specifying the desired level of accuracy~\cite{30}. The study in~\cite{28} suggested to employ the time series specific classifiers to develop an efficient cost function for early classification.
\subsubsection{\textbf{Miscellaneous approaches}} A dynamic adjustment strategy is recommended in~\cite{31, 49} for setting the reward function parameters in the reinforcement learning-based early classification approach. The studies in~\cite{16, 21} suggested to explore domain-dependent density functions to capture the structure of time series data. A hybrid model can be developed to incorporate statistical features with the incomplete time series during its early classification~\cite{53}. The work in~\cite{22} recommended to improve the interpretability of the neurons in neural networks for addressing the imbalanced distribution of the training instances.

\bibliographystyle{IEEEtran}
\bibliography{reference}

\begin{IEEEbiography}  
[{\includegraphics[height=1in, width = 1in]{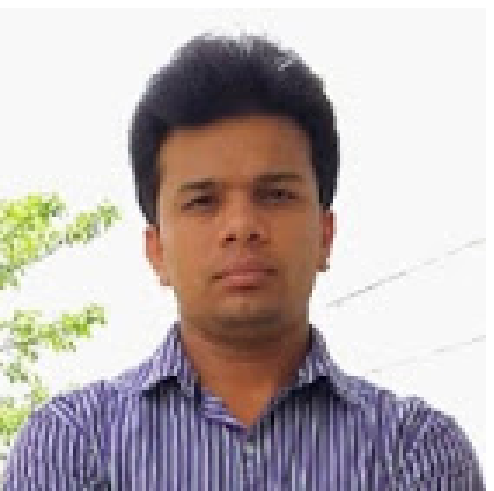}}] 
{~\\Ashish Gupta} received the M.Tech. degree in Computer Engineering from Sardar Vallabhbhai National Institute of Technology, Surat, India, in 2012. Presently, he is pursuing Ph.D. in Computer Science and Engineering, Indian Institute of Technology (BHU) Varanasi, India. His research interests include data analytics, IoT, data structures and algorithms, and machine learning.
\end{IEEEbiography}

\begin{IEEEbiography}  
[{\includegraphics[height=1in, width = 1in]{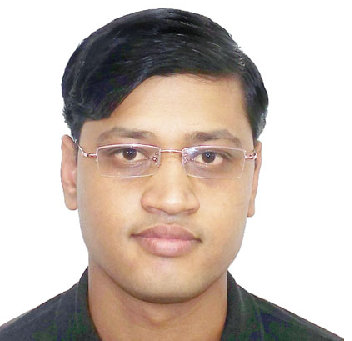}}] 
{~\\Hari Prabhat Gupta} received the Ph.D. degree in Computer Science and Engineering from the Indian Institute of Technology Guwahati, India, in 2014. He is currently working as Assistant Professor in Department of Computer Science and Engineering, Indian Institute of Technology (BHU), Varanasi, India. His research interests include wireless sensor networks, distributed algorithms, and IoT.
\end{IEEEbiography}

\begin{IEEEbiography}  
[{\includegraphics[height=1in,width = 1in]{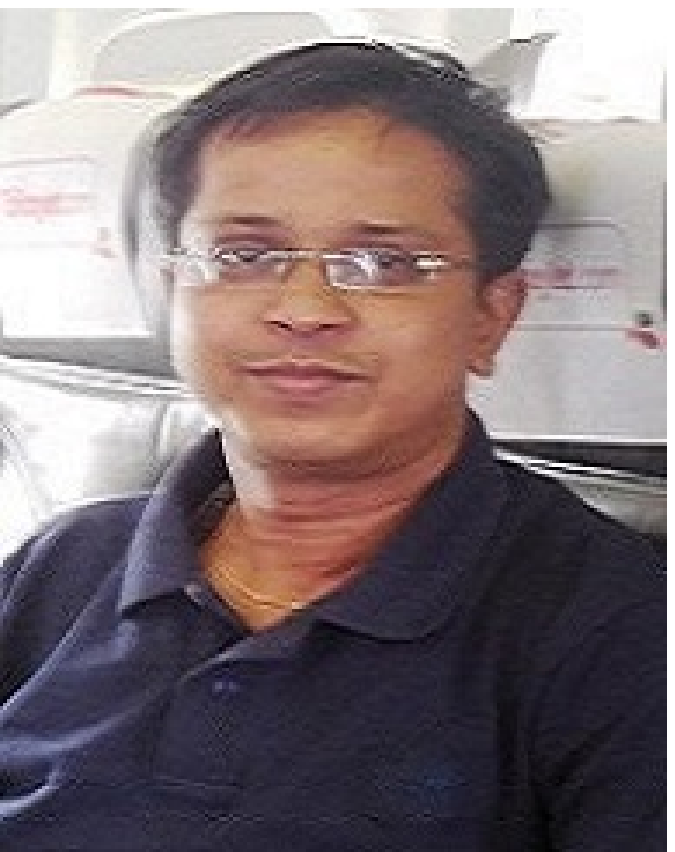}}] 
{~\\Bhaskar Biswas} received the Ph.D. in Computer Science and Engineering from Indian Institute of Technology (BHU), Varanasi. He is working as Associate Professor at Indian Institute of Technology (BHU), Varanasi
in the Computer Science and Engineering department. His research interests include data Mining, text analysis, online social network analysis, and evolutionary computation.
\end{IEEEbiography}

\begin{IEEEbiography}  
[{\includegraphics[height=1in,width = 1in]{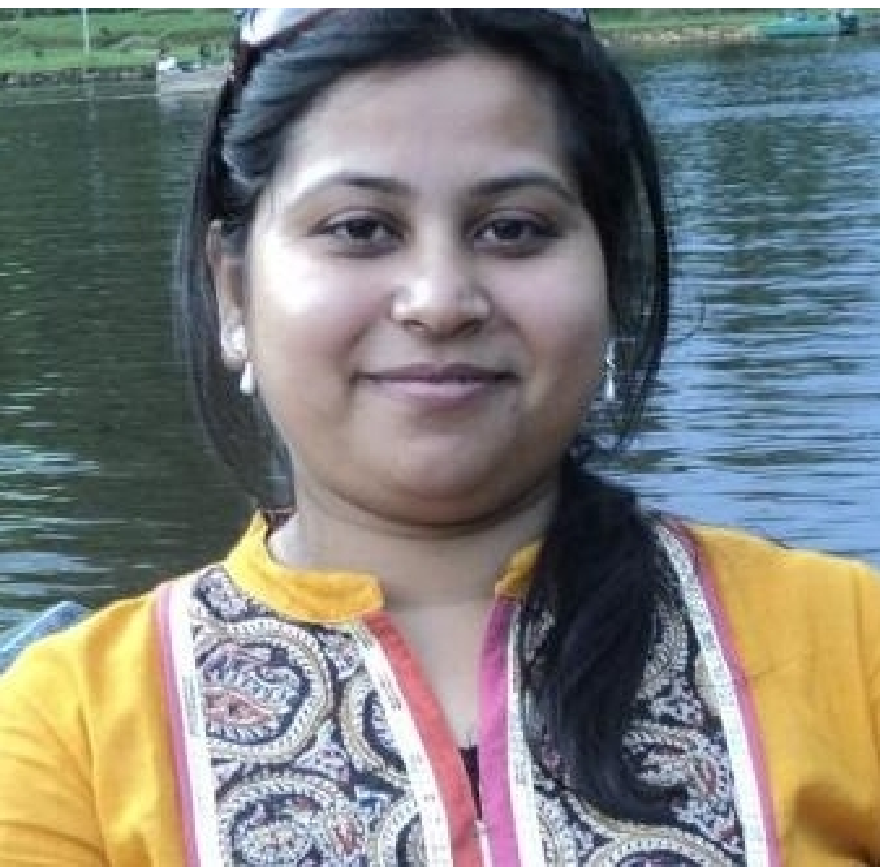}}] 
{~\\Tanima Dutta} received the Ph.D. degree in computer science and engineering from Indian Institute of Technology Guwahati, India, in 2014. She is currently working as Assistant Professor in Department of Computer Science and Engineering, Indian Institute of Technology (BHU), Varanasi, India. Her research interests include multimedia forensics, deep learning, and human computer interaction.
\end{IEEEbiography}

\end{document}